\documentclass[conference]{IEEEtran}
\IEEEoverridecommandlockouts
\usepackage{cite}
\usepackage{amsmath,amssymb,amsfonts}
\usepackage{algorithmic}
\usepackage{graphicx}
\usepackage{textcomp}
\usepackage{xcolor}
\def\BibTeX{{\rm B\kern-.05em{\sc i\kern-.025em b}\kern-.08em
    T\kern-.1667em\lower.7ex\hbox{E}\kern-.125emX}}

\newcommand{\R}{\mathbb{R}}

\newcommand{\norm}[1]{\left\lVert#1\right\rVert}
\DeclareMathOperator{\Tr}{Tr} 

\newcommand{\entity}[1]{\mbox{\textsf{#1}}}

\newcommand{\emb}{\mathbf}

\newcommand{\noshow}[1]{}
    
\begin{document}

\title{Repurposing Knowledge Graph Embeddings for Triple Representation via Weak Supervision}

\author{\IEEEauthorblockN{1\textsuperscript{st} Alexander Kalinowski}
\IEEEauthorblockA{\textit{Department of Information Science} \\
\textit{Drexel University}\\
Philadelphia, USA \\
ajk437@drexel.edu}
\and
\IEEEauthorblockN{2\textsuperscript{nd} Yuan An}
\IEEEauthorblockA{\textit{Department of Information Science} \\
\textit{Drexel University}\\
Philadelphia, USA \\
ya45@drexel.edu}}

\maketitle

\begin{abstract}
The majority of knowledge graph embedding techniques treat entities and predicates as separate embedding matrices, using aggregation functions to build a representation of the input triple. 
However, these aggregations are lossy, i.e. they do not capture the semantics of the original triples, such as information contained in the predicates.
To combat these shortcomings, current methods learn triple embeddings from scratch without utilizing entity and predicate embeddings from pre-trained models. 
In this paper, we design a novel fine-tuning approach for learning triple embeddings by creating weak supervision signals from pre-trained knowledge graph embeddings. 
We develop a method for automatically sampling triples from a knowledge graph and estimating their pairwise similarities from pre-trained embedding models.
These pairwise similarity scores are then fed to a Siamese-like neural architecture to fine-tune triple representations.
We evaluate the proposed method on two widely studied knowledge graphs and show consistent improvement over other state-of-the-art triple embedding methods on triple classification and triple clustering tasks.
\end{abstract}


\section{Introduction}
\label{sec:introduction}

Knowledge graphs (KG) are a paradigm for representing head entities (subjects), tail entities (objects) and the predicates between them ($\langle h,p,t \rangle$ triples) and form a computational-basis for the Semantic Web.
Due to their size and incompleteness, many approaches have been developed for compressed representations of KGs, with the majority of these approaches relying on low-dimensional vector representations, henceforth referred to as \emph{embeddings}.
These embeddings have been shown to be highly effective in link prediction and knowledge graph completion tasks.
However, almost all methods encode entities and predicates in their own independent embedding vectors, requiring additional functions to combine the resulting vectors into a single \textit{`triple'} representation.
Such \textit{`triple'} embeddings have applications in many downstream tasks, including relation extraction~\cite{kalinowski-sentence-structure} and e-commerce predictions~\cite{triple2vec}.

Functions explored in the literature typically combine the head and tail entity representations for a composite \textit{`triple'} representation, with good performance on simple, homogeneous graphs.
Knowledge graphs, on the other hand, are heterogeneous multi-graphs where each predicate edge is associated with a particular semantic label, thus causing issues with the traditional methods of aggregation for triple representation.


In a heterogeneous multi-graph, there may exist multiple different predicates between two entities.
For example, a person, 
\entity{David Lynch}, can play the role of both an actor and a director in a particular piece of work, in this case the television series \entity{Twin Peaks}:
\entity{"David Lynch"} -- \entity{actor\_in} -- \entity{"Twin Peaks"}, and 
\entity{"David Lynch"} -- \entity{director\_of} -- \entity{"Twin Peaks"}. 
Triple representations that rely only on an aggregation of the embeddings of the head and tail entities miss the contextual information contained in the multitude of predicates.
An alternate approach for encoding a triple is to include the embeddings of all three components of the triple.
However, aggregations of all three components may offset the information encoded in the entity and predicate embeddings. 
For example, a main category of KG embedding approaches is translation-based models such as TransE~\cite{transe}.
These models assume that predicates can be modelled as translations from the head entity to the tail entity, $\emb{h} + \emb{p} \approx \emb{t}$.
When combining these representations into a triple, a simple mechanism would be to sum or average all component vectors that comprise the triple, i.e. 
$
	\emb{e}_{triple} = \emb{h} + \emb{p} + \emb{t}
	\label{eqn:add-trip}
$
where $\emb{e}_{triple}$ represents the embedding vector of the entire triple $\langle h, p, t\rangle$.
Clearly, this is problematic given the assumption that $\emb{h}$ $+$ $\emb{p}$ $\approx$ $\emb{t}$, substitution into the equation~(\ref{eqn:add-trip}) would lead to a representation of the triple that is twice the embedding of the tail entity, i.e.
$$\emb{e}_{triple} = \emb{h} + \emb{p} + \emb{t} \approx \emb{t} + \emb{t} = 2\emb{t}$$
where the relational semantics of the triple, including any information about the predicate involved, are entirely lost.

To treat a knowledge graph triple as a first-class object and represent it as a single vector, Triple2vec ~\cite{triple2vec} learns triple embeddings completely from scratch, based on the idea of random walks on the line graph of the original graph.
Building new triple representations from scratch ignores the information already captured in embedding algorithms designed for link prediction.
In contrast, we believe that knowledge graph embedding methods still maintain their merits, yet require re-working to craft a triple representation.
Rather than building these triple embeddings from scratch, we aim at re-usability and flexibility.
Following suit with the works detailed in~\cite{word-pp}, we build a post-processing and fine-tuning technique to re-use knowledge graph embeddings, saving on computational cost, especially in the case where pre-trained models are readily available~\cite{libkge}.

\begin{figure*}
	\centering
	\includegraphics[width=.85\textwidth]{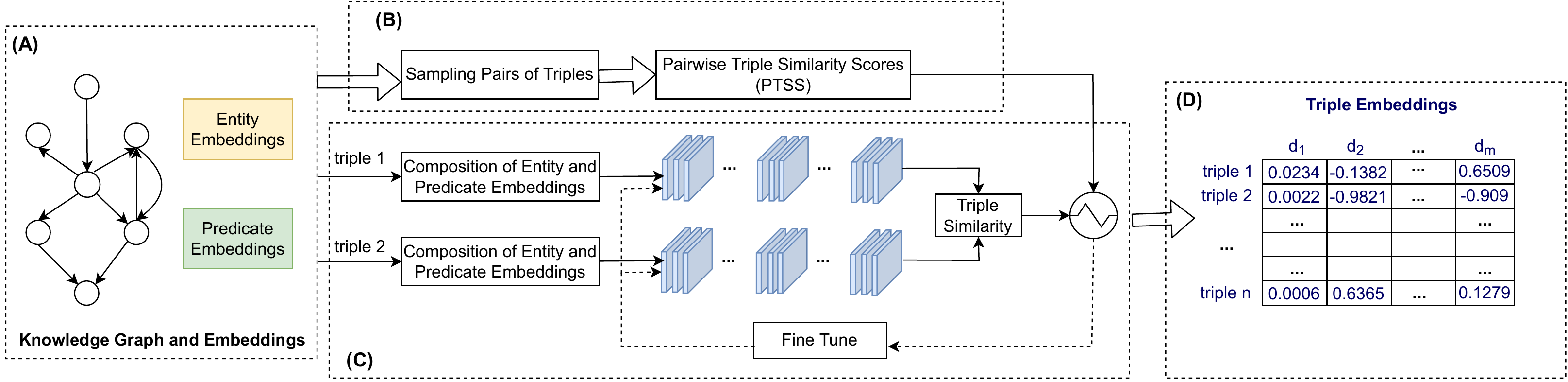}
	\caption{
	The architecture of the weakly supervised learning system for triple embeddings. 
	(A) The input of the system consists of a knowledge graph and its pre-trained entity and predicate embeddings. 
	(B) A sampling method automatically draws pairs of triples for generating pairwise triple similarity scores (PTSS). 
	(C) A Siamiese-like neural network takes a pair of triples and composes an initial triple embedding using the constituent embeddings.  
    Initial embeddings are forward propagated through the network with the resulting outputs scored by a similarity function. 
    Error between this score and the PTSS is used to refine the parameters of the entire network.
    (D) After training, the system is able to generate embeddings for all triples.}
	\label{fig:architecture}
\end{figure*}

The key idea of our method is to learn triple embeddings by post-processing pre-trained entity and predicate embeddings in a \emph{weakly supervised manner}. 
Our main inspiration comes from the ideas of semantic textual similarity~\cite{sts-benchmark} and Siamese neural networks~\cite{sentencebert}. 
Figure \ref{fig:architecture} illustrates the architecture of our system. 
The method consists of two main components (labeled as (B) and (C) in Figure \ref{fig:architecture}). 
In the first component (B), we propose a novel method for generating weak supervision signals by combining pre-trained entity and predicate embeddings (from (A)) in a semantic way.
\noshow{namely, triples sharing similar elements should have similar low-dimensional representations.}
In the second component (C), we apply these signals to fine-tune triple embeddings composed by a Siamese-like deep neural network. 

Following the pioneering work in Triple2vec~\cite{triple2vec} on evaluating triple embeddings, we also used the tasks of triple classification and triple clustering and compared their approach to our new proposed methodology.
When evaluated on two widely studied knowledge graphs, our method consistently surpasses all baselines.

The rest of the paper is organized as follows. 
Section \ref{sec:formulation} presents preliminaries and formalizes the problem. 
Section \ref{sec:building-triple-similarity-scores} describes the process of sampling triples and building pairwise triple similarity scores.
Section \ref{sec:learning-triple-embeddings} presents the weakly supervised method for learning triple embeddings. 
Section \ref{sec:evaluation} describes the evaluation settings and process. 
Section \ref{sec:analysis-results} analyzes the evaluation results in comparison to baselines. 
Section \ref{sec:related-work} discusses related work. 
Finally, Section \ref{sec:conclusion} concludes the paper with ideas on future improvements.

%
%

%
%
\section{Problem Formulation}
\label{sec:formulation}


A knowledge graph is a 3-tuple $G=(E, P, T)$, where $E$ is a set of entities, $P$ is a set of predicates, and $T$ is a set of triples $\langle h, p, t\rangle$ connecting a head entity $h\in E$ to a tail entity $t\in E$ by a predicate $p\in P$ as an edge. 
A knowledge graph embedding (KGE) method is function 
$f_{e}: E\rightarrow \R^d$ which encodes entities $v\in E$ as $d$-dimensional vectors $\emb{e}_{v}\in \R^{d}$ and predicates $r\in P$ as a scoring functions $f_{r}: \R^{d}\times \R^{d} \rightarrow \R$.

The \emph{PTSS triple embedding problem} is to find a function $g: T \rightarrow \R^{d}$ such that 
$g$ reuses the output of KGE functions to encode each triple as a $d$-dimensional vector. 
In this paper, we implement $g$ via weak supervision signals from KGE similarity scores.


%
%
\section{Building Triple Similarity Scores}
\label{sec:building-triple-similarity-scores}

We begin by building weak supervision signals for learning triple embeddings. 
We seek signals corresponding to similarity measures between the triples that share a single element (somewhat similar), those that share multiple elements (more similar), and those that do not share any elements (not similar). 
Unfortunately, knowledge graph benchmark datasets (i.e. WordNet or FreeBase) do not have explicit measures of the similarity between two triples.
Building these similarity scores could require subjective human labelling, usually accomplished through crowdsourcing, as is done with semantic textual similarity (STS) tasks~\cite{sts-benchmark}.
Rather than trust human judgment of these similarities, we build a scoring mechanism based on the low-dimensional, pre-trained embeddings of the entities and predicates that compose each triple. 

Let $G$ be a KG and $f$ a corresponding KGE function.
For triples $\langle h, p, t \rangle$ with $h,t \in E$ and $p \in P$, the embeddings learned through $f$ may already encode vector space notions of similarities between entities and predicates.
For example, two entities sharing many properties and relationships will participate in many of the same triples, and we assume that their respective embeddings should exhibit these similarities. 
This assumption is highly dependent on the selection of the pre-trained embedding function $f$, and there exists a line of research into how much semantic information these embeddings actually encode~\cite{jain2021do}. 
We can use these embeddings, henceforth referred to as \textit{seed embeddings}, as a means to proxy pairwise triple similarity scores.
Specifically, given two triples $T_a = \langle h_a, p_a, t_a \rangle$ and $T_b = \langle h_b, p_b, t_b \rangle$ in the knowledge graph, we define the \textit{pairwise triple similarity score (PTSS)} between $T_a$ and $T_b$ as an average of the similarities between the embeddings of their head entities, tail entities, and predicates, respectively. 
Formally,
\begin{center}
	\label{eq:ptss}
	$PTSS(T_a, T_b) = avg(sim(\emb{e}_{h_a}, \emb{e}_{h_b}), sim(\emb{e}_{p_a}, \emb{e}_{p_b}), sim(\emb{e}_{t_a}, \emb{e}_{t_b}))$,
\end{center}
where $\emb{e}_{h_a}$ is the embedding vector of the head entity $h_a$ and so on, $avg(x, y, z)$ computes an average of its arguments, and the function $sim(\emb{e}_{1}, \emb{e}_2)$ is a function for computing the similarities of two embeddings. 
The similarity function $sim(\emb{e}_{1}, \emb{e}_2)$ could be an arbitrary function capturing the geometric structures in KG embedding spaces. 
In our work, we choose cosine similarity and arithmetic mean, leaving the evaluation of more general functions to future work. 
Our evaluation demonstrates that these simple functions generate useful weak supervision signals. 

For the proposed weakly supervised training, we extract from the knowledge graph a set of training examples containing both \emph{"positive"} and \emph{"negative"} examples.
For a triple $T_a = \langle h_a, p_a, t_a \rangle$, we define positive training examples as a set of potential \textit{candidate matches}.
In particular, triple $T_a$ is composed of three elements, or three potential `slots' to match on: the head, the tail and the predicate.
For each `slot' we identify candidate matches as follows.
Select $N$ other triples with the same head entity, $N$ other triples with the same tail entity, $N$ other triples with the same predicate.
For negative examples, we select $N$ other triples with no commonalities in any available slot.
In instances where the set of candidate matches has cardinality less than $N$, we add the entire set to be scored.
Thus, for each triple in the graph, we build a set of at most $4N$ triples to compare and contrast with.

These scores and training examples can then be utilized as an input to a fine-tuning process to convert the low-dimensional representations of entities and predicates into low-dimensional representations of the composite triple.
For any arbitrary knowledge graph, we can build these similarity estimates in an automated way given that we specify an initial embedding function $f$, a similarity function $sim(\emb{e}_1, \emb{e}_2)$, an averaging function $avg(x, y, z)$, and a sampling parameter $N$.
We note that these scores are not meant to be gold standard labels; instead, they are utilized as weak supervision signals to bootstrap the triple similarity training process.

%
%
\section{Learning Triple Embeddings}
\label{sec:learning-triple-embeddings}

For learning triple embeddings, we define a Siamese-like neural network that takes two triples as input and initializes values at an embedding layer.  
The network subsequently forward propagates the values at the embedding layer through a series of encoding (ENC) layers.
The final layer is a scoring (SCO) layer where the network computes the cosine similarity of the representations.
This score is then compared to the estimated pairwise similarity scores (PTSS) defined in the previous section, with the errors back-propagated through the network. 
To update the triple embeddings, we make the embedding layer tunable and use the final values as the triple embeddings.
In many neural network approaches, embedding layers are typically initialized randomly or following the Xavier normalization scheme~\cite{xavier}.
In our approach, we are leveraging pre-trained knowledge graph embeddings, and thus use this information as a starting point for the initialization of this layer.
We experiment with five aggregation methods for combining the embeddings of the head and tail entities into a triple representation.
Four of these aggregation methods are utilized in~\cite{node2vec}; here, we introduce a fifth: the concatenation of the head and tail vectors. 
These choices are outlined in Table~\ref{tab:agg-methods}.
We investigate the impacts of these aggregation functions to the learned triple embeddings in conjunction with the selected underlying KGE methods. 
The results provide insights for adopting or generalizing our methods in future applications.  

\begin{table}
\caption{\label{tab:agg-methods}A summary of aggregation operators for constructing triple representations}
\center
\begin{tabular}{lc}
\hline
Operator & Definition \\
\hline
Average (AVG) &  $[f(\emb{u}) \star f(\emb{v})]_i = \frac{f_i(\emb{u}) + f_i(\emb{v})}{2}$ \\
Hadamard (HAD) & $[f(\emb{u}) \star f(\emb{v})]_i = f_i(\emb{u}) * f_i(\emb{v})$\\
Weighted-L1 (L1) & $| f_i(\emb{u}) - f_i(\emb{v}) |$ \\
Weighted-L2 (L2) & $| f_i(\emb{u}) - f_i(\emb{v}) |^2$ \\
Concatenation (HT) & $[f(\emb{u}) \star f(\emb{v})]_i = f_i(\emb{u}) \| f_i(\emb{v})$\\
\hline
\multicolumn{2}{l}{\footnotesize \textbf{Notes}:}\\
\multicolumn{2}{l}{\footnotesize `$*$' denotes element-wise product operation.} \\
\multicolumn{2}{l}{\footnotesize `$\|$' denotes vector concatenation operation.}
\end{tabular}
\end{table}

Formally, an encoding (ENC) and scoring (SCO) layer
of the network is defined as
\begin{center}
$ \emb{o}_a = \sigma(\emb{W}_1(\emb{e}_{T_a}) + \emb{b}_1) $ \\
$ \emb{o}_b = \sigma(\emb{W}_1(\emb{e}_{T_b}) + \emb{b}_1) $ \\
$ \hat{s} = \cos(\emb{o}_a, \emb{o}_b) $ \\
$ \mathcal{L} = MSE(s_{a,b}, \hat{s})$
\end{center}
where $\emb{e}_{T_a}$ and $\emb{e}_{T_b}$ are embeddings of two triples $T_a$ and $T_b$, $\emb{W}_1$ is a shared dense linear layer, $\emb{b}_1$ the bias of that layer, and $\sigma$ a non-linearity, in this case hyperbolic tangent.
The outputs $\emb{o}_1$ and $\emb{o}_2$ are then compared using cosine similarity to produce a final estimate of the similarity between the two inputs, $\hat{s}$.
The loss is defined as the mean squared error between the computed similarity $\hat{s}$ and the estimated pairwise triple similarity score $s_{a,b} = PTSS(T_a, T_b)$ defined in Section \ref{sec:building-triple-similarity-scores}.
For our experiments, we use a single encoding and scoring layer (ENC-SCO layer) to measure the efficacy of the weakly supervised approach.
We note that many of these layers could be stacked to arbitrary depth. 
Our evaluation shows that a single layer can lead to significant improvements over the baselines and aligns with our goal of re-usability in compute-restricted environments. 
We leave the evaluation of more complex network structures for future applications. 

%
%
\section{Evaluation}
\label{sec:evaluation}

\subsection{Data Sets Selection}
\label{subsec:data-sets}

In line with Triple2vec~\cite{triple2vec}, we construct two tasks, \textit{triple classification} and \textit{triple clustering}, to evaluate our learned embeddings.
However, we deviate from their work in selection of our benchmark datasets and utilize two well-established knowledge graphs, WordNet (WN18RR) and Freebase (FB15K-237).
Summary statistics for both datasets are presented in Table~\ref{tab:datasets}.
These benchmark datasets were selected for two reasons.
First, FB15K-237 and WN18RR are utilized as a benchmark for all of the seed knowledge graph embedding methods we tested.
The frequency of usage in the research community also allows us to take advantage of publicly available, pre-trained models, such as those found in LibKGE~\cite{libkge}.
Secondly, these graphs are complex in nature, making our evaluation more reflective of knowledge graphs that occur in practice.

\begin{table}
\caption{\label{tab:datasets} Summary of Benchmarks}
\centering
\begin{tabular}{lcc}
\hline
                 Dataset &    WN18RR &    FB15K-237   \\
\hline
                 Entities & 40,943 & 14,541 \\
                 Predicates & 11 & 237 \\
                 Triples & 93,003 & 310,116 \\
                 Multi-edge Triples & 218 & 49,214\\ 
                 Num. Strongly CC & 23,105 & 2,678 \\
                 Num. Weakly CC & 46 & 6 \\
                 
\hline
\end{tabular}
\end{table}

Important to our discussion is the underlying topology of these graphs.
We note that neither graph is connected, meaning that a random walker on each graph may fail to find a path between two arbitrarily selected nodes.
Furthermore, the distributions of strongly and weakly connected components are very different.
WN18RR presents itself as a denser graph: more strongly and weakly connected components as compared to FB15K-237, which is sparser and has only six weakly connected components.
The differences in the graph topologies are an important factor for us to consider in the analysis of our results, especially in comparison to the Triple2vec approach which is dependent on random walks over the graph.
Of particular interest to our approach is the number of triples that contain multiple predicates between the head and tail entities.
In WN18RR, this occurs very infrequently; the overwhelming majority of triples contain a single predicate.
On the other hand, 16\% of triples found in FB15K-237 contain more than one predicate.
From this difference, we would expect our approach to have a greater impact on triple classification metrics when applied to FB15K-237 over WN18RR.

\subsection{Triple Classification Task}
\label{subsec:triple-classification-task}

To test the efficacy of our proposed model, we probe the resulting triple embeddings for their ability to predict their respective predicate labels.
Following the work of~\cite{senteval} and ~\cite{triple2vec}, we perform this classification using both a low-capacity and high-capacity model.
For our low-capacity model, we employ one-vs-rest logistic regression, providing the model with triple representations as features and the predicate indices as labels.
For the high-capacity model, we replace the logistic regression with a multi-layer perceptron with 512 hidden nodes, trained with the Adam optimizer and batch size of 256 for a total of 10 epochs.
In each case, we split the triples into training and test sets, with the training set comprising of 80\% of the total triples with the remaining 20\% for testing, repeated using five non-overlapping folds.
Both models are then evaluated on their respective Micro-F1 scores.

\subsection{Triple Clustering Task}
\label{subsec:triple-clustering-task}

Motivated by work on word embedding regularities~\cite{exploit-sim}, we wish to probe the triple embedding spaces generated in our experiments to measure the degree to which they exhibit an underlying structure.
Our hypothesis is that embedding techniques that exhibit higher degrees of clusterability are able to capture more of the semantics expressed in the input triples.
To formalize this notion, we introduce a definition of clusterability~\cite{clusterability}.
For some dataset $X \subseteq \mathbb{R}^n$, a description of the clusterability of $X$ is a function $c: X \rightarrow v$ where $v \in \mathbb{R}$ is a real value.
Here, $v$ is a measure of how strong a clustering presence is in the underlying set $X$.
As t-SNE plots, such as those in the clustering evaluation of~\cite{triple2vec}, can be difficult to tune and interpret~\cite{effective-tsne}, we choose to evaluate clusterability via a metric, although code for generating these plots is included in our open source implementation.
Our process for measuring embedding space clusterability is as follows.
For each triple embedding space, we apply k-means clustering, selecting k to be the number of known predicates in the graph.
To measure the quality of these clusters, we select the Calinski-Harabasz (CH) index~\cite{ch-metric}, which is the ratio of between cluster dispersion to within cluster dispersion.
Mathematically, this score is computed as 
$$ v = \frac{\Tr(B_k)}{\Tr(W_k)} \times \frac{n_T - k}{k - 1}, $$
where $n_T$ is the number of triples, $k$ the number of clusters, $\Tr(B_k)$ the trace of the between cluster dispersion matrix and $\Tr(W_k)$ the trace of the within cluster dispersion matrix.
Higher values of the CH metric indicate a better fit of the k-means partitioning. 
This metric has an added benefit in that it tends to be inflated for convex clusters; we hypothesize convex structures will allow for well-defined areas corresponding to the predicates we seek to predict.

\subsection{Experimental Design}
\label{subsec:experimental-design}

We choose several popular knowledge graph embedding models for seed embeddings,
including \textbf{TransE}, \textbf{DistMult}, \textbf{RESCAL}, \textbf{ConvE}, \textbf{RotatE} and \textbf{ComplEx} (See Section \ref{sec:related-work} for more details).
We leverage the open source LibKGE package and associated pre-trained models~\cite{libkge} for their implementations.
These pre-trained models have undergone extensive hyperparameter tuning to achieve the highest link prediction accuracies.
For comparison with \textbf{Triple2vec}, we use an implementation provided by the authors~\cite{triple2vec}, extended to our benchmark knowledge graphs and trained for 30 epochs.
To minimize the impact of the choice of embedding dimensionality on the triple classification task, we train Triple2vec with dimension matching the dimension of pre-trained models from our approach and set the hyperparameter for the number of walks to be 10.

For all of our experiments, we set the PTSS hyperparameter $N$ to be 5, selected randomly from the set of all candidate matches, although we note that this hyperparameter deserves further investigation and tuning to measure its full effect.
We depend on the embedding dimension of each pre-trained knowledge graph model to be the input dimension of each triple embedding space.
For the remainder of model training, we set the batch size to 128, optimize using Adam with an initial learning rate of 2e-3 and warm up each model using 10 percent of the training data.



\subsection{Implementation and Baselines}
\label{subsec:baselines}
We dub our proposed method as PTSS in the evaluation. The source code implementation is 
available here\footnote{https://github.com/yur7nd/ptss}. The baselines to which we compare our method are (1) the state-of-the-art triple embeddings method, Triple2vec~\cite{triple2vec} and (2) aggregated pre-trained KG embeddings by applying the operators in Table \ref{tab:agg-methods} without our PTSS fine-tuning.

%
%

\begin{figure}
	\centering
	\includegraphics[width=.5\textwidth]{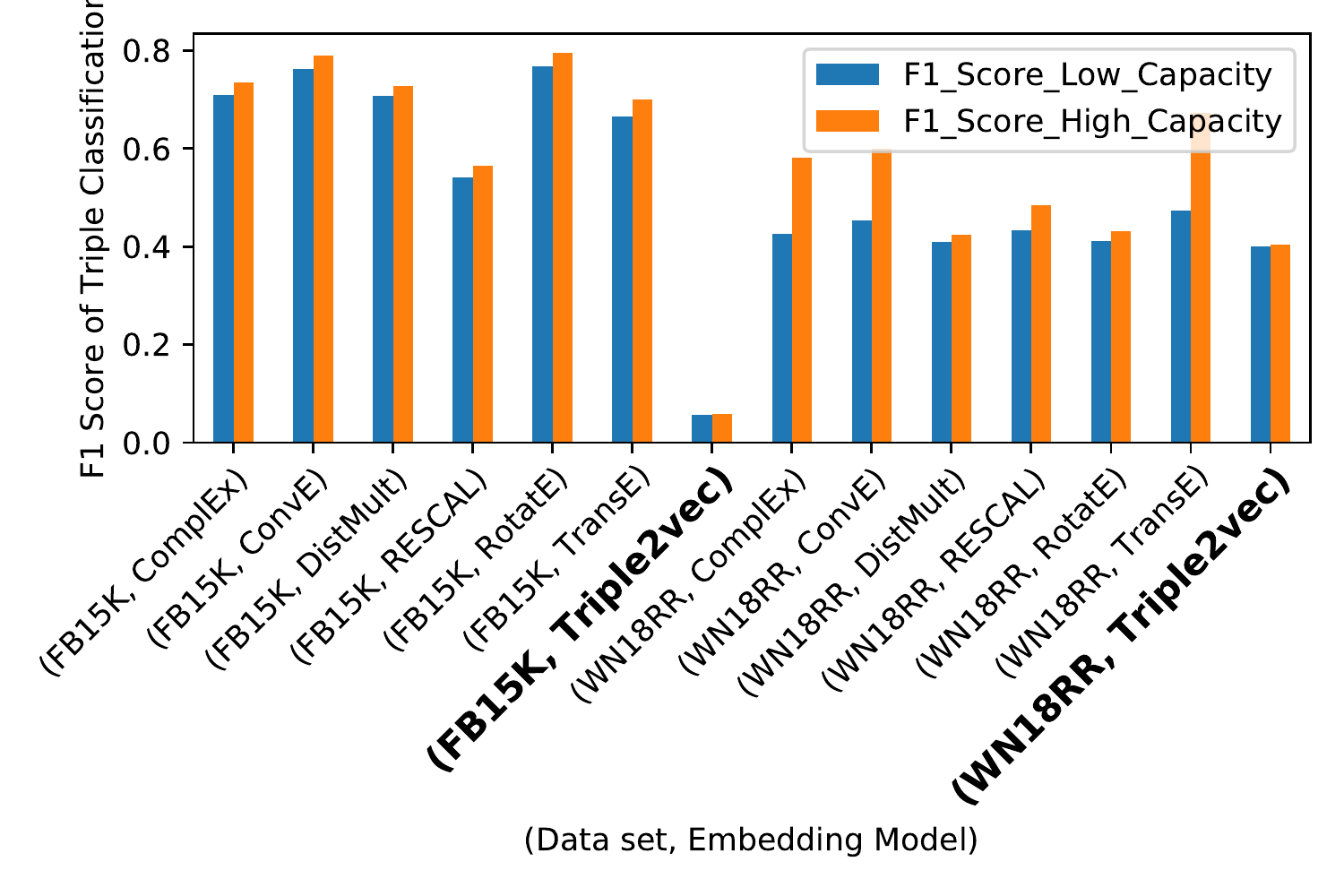}
	\caption{Comparison of Micro-F1 scores for Triple2vec and PTSS for the triple classification task. For display purposes, we selected the best performing model/aggregation combination.}
	\label{fig:all_comparison}
\end{figure}

\section{Analysis of Results}
\label{sec:analysis-results}

\subsection{Comparison of Triple2vec and PTSS}
Overall, we find that the vectors generated by PTSS serve as better features for triple classification than those generated from Triple2vec. 
Figure~\ref{fig:all_comparison} show the comparison of Micro-F1 scores between Triple2vec and the best performing model and aggregation combinations from our PTSS approach.
On WN18RR, our low-capacity models perform better than, but on par with, the results of Triple2vec, while our triple vectors serve as better features for the high-capacity classifier.
The best model, a combination of TransE seed vectors and weighted-L2 aggregation, has F1 score of 0.6723, while Triple2vec representations with equal dimension only achieves F1 score of 0.4008.
The performance gains are even more dramatic on FB15K-237.
Our best model (RotatE seed vectors with aggregation by concatenation) achieves F1 score of 0.7949 while the best Triple2vec representation of equal dimension only reaches 0.059.
The rationale for such a gap in performance on FB15K-237 comes from the topology of the graph; fewer connected components and lower degree means random walks are more likely to reach dead ends.
Triple2vec was inherently designed to take advantage of these random walkers; hence, when the graph is less densely connected, fewer pieces of information are available for their approach to learn from.
Our approach is not dependent on random walks and is invariant to the underlying graphs topology.
In comparison to Triple2vec, we also note that our approach is more computationally efficient.
Building the line graph $G_L$ from the input knowledge graph $G$ can be burdensome, with a cost of $\mathcal{O}(|T|^2 + costWeight)$, where $costWeight$ is a factor for computing edge weights (see~\ref{sec:graph-models}).
In comparison, building the PTSS dataset can be accomplished in $\mathcal{O}(4N\times |T|)$ where $N \ll |T|$.

\subsection{Assessing the Aggregation Operator}

We analyze the impact of the aggregation operations on the triple classification performance.
We find that there is no clear aggregation function that consistently outperforms all others. 
For WN18RR, two models showed best performance using the Hadamard product (ComplEx and ConvE), two models with top performance using weighted-L1 (DistMult and TransE), while RESCAL and RotatE performed best with the concatenation and average operators, respectively.
On FB15K-237, ComplEx and DistMult performed best with the Hadamard product, ConvE and RESCAL with weighted-L2 and RotatE and TransE with the concatenation operator.
These results are counter to our intuition; our expectation, aligned with results of other lines of research~\cite{node2vec}, was that Hadamard product would consistently outperform all other operators.
It was interesting to note that the performance of these operators with respect to the selected embedding model was also inconsistent across datasets.
Only the ComplEx model showed best performance when combined with the Hadamard product for both datasets.
We also note that while simple, the concatenation method does yield positive results for three of our tests: in conjunction with RESCAL on WN18RR and with RotatE and TransE on FB15K-237.
We expected this simple concatenation to benefit from a performance boost due to the extra triple feature dimensionality (twice that of all other approaches due to the nature of vector concatenation), yet we do not see a consistent boost across the board.
This is further explained by the correlation of triple feature dimensionality with model performance, which has a neutral correlation of 0.143.
Given the lack of concrete evidence suggesting that any aggregation operator is superior, 
we recommend treating this choice as a hyperparameter to be tuned when applying our methodology to novel knowledge graphs.

\subsection{Quality of Input Embeddings}

We compare the quality of the input seed embeddings to the results of the triple classification task.
As the input embeddings are all trained for the link prediction task, we present their reported MRR, Hits@1, Hits@3 and Hits@10 from~\cite{libkge} and assess the correlation of these metrics with our reported Micro-F1 scores on triple classification.
Actual performance metrics from~\cite{libkge} are reproduced with the best triple classification Micro-F1 scores in Tables~\ref{tab:lp-results-fb} and ~\ref{tab:lp-results-wn}.

\begin{table}
\caption{Link Prediction Performance on FB15K-237}
\centering
\begin{tabular}{lccccc}
\hline
                 Approach & MRR & Hits@1 & Hits@3 & Hits@5 & Best F1* \\
                 \hline
                 ComplEx & 0.348 & 0.253 & 0.384 & 0.536 & 0.7346 \\
                 ConvE & 0.339 & 0.248 & 0.369 & 0.521 & 0.7896 \\
                 DistMult & 0.343 & 0.250 & 0.378 & 0.531 & 0.7268 \\
                 RESCAL & \textbf{0.356} & \textbf{0.263} & \textbf{0.393} & \textbf{0.541} & 0.0657 \\
                 RotatE & 0.333 & 0.240 & 0.368 & 0.522 & \textbf{0.7949} \\
                 TransE & 0.313 & 0.221 & 0.347 & 0.497 & 0.7008 \\
\hline
\multicolumn{6}{l}{\footnotesize *Best F1: the best Micro-F1 scores of our method on triple classification.}
\end{tabular}
\label{tab:lp-results-fb} 
\end{table}

\begin{table}
\caption{Link Prediction Performance on WN18RR}
\centering
\begin{tabular}{lccccc}
\hline
                 Approach & MRR & Hits@1 & Hits@3 & Hits@5 & Best F1* \\
                 \hline
                 ComplEx & 0.475 & 0.438 & 0.490 & 0.547 & 0.582 \\
                 ConvE & 0.442 & 0.411 & 0.451 & 0.504 & 0.600 \\
                 DistMult & 0.452 & 0.413 & 0.466 & 0.530 & 0.424 \\
                 RESCAL & 0.467 & \textbf{0.439} & 0.480 & 0.517 & 0.484 \\
                 RotatE & \textbf{0.478} & \textbf{0.439} & \textbf{0.494} & \textbf{0.553} & 0.431\\
                 TransE & 0.228 & 0.053 & 0.368 & 0.520 & \textbf{0.672} \\
\hline
\multicolumn{6}{l}{\footnotesize *Best F1: the best Micro-F1 scores of our method on triple classification.}
\end{tabular}
\label{tab:lp-results-wn}
\end{table}

We find little evidence of consistent, strong positive correlations between performance on the link prediction task and performance on the triple classification task ($\rho =-0.6377$ and $\rho =0.4285$ for FB15K-237 and WN18RR, respectively).
It is important to note that the best performance on the link prediction task does not imply best performance on triple classification.
Our architecture is not merely a pass-through mechanism--we are able to add semantics back to triple representations that were previously lost in the pre-trained embedding models.
It is also interesting to note that poor performing models on the link prediction task can still be re-purposed to yield strong triple embeddings.
On WN18RR, the TransE approach has the worst performance for link prediction, but produces entity and predicate embeddings that can be best used for the pairwise triple similarity scoring required by our model.
For the more complex data in FB15K-237, increasing the complexity of the pre-trained embeddings does help performance on the triple classification task--more complicated models such as RotatE and ConvE are able to achieve superior performance. 
As with the impact of the aggregation operator, the choice of which model used to seed the entity and predicate embeddings entirely depends on the underlying dataset; there is no silver bullet that always achieves superior performance.
The inability to find a clear, consistent winner across all our experiments further highlights an important feature of our approach: flexibility.
Our methodology is general enough to be adapted to different aggregation operators and sets of pre-trained embeddings, all dependent on the complexity of the knowledge graph being investigated.

\subsection{Ablation of PTSS}

\begin{figure}
	\centering
	\includegraphics[width=.5\textwidth]{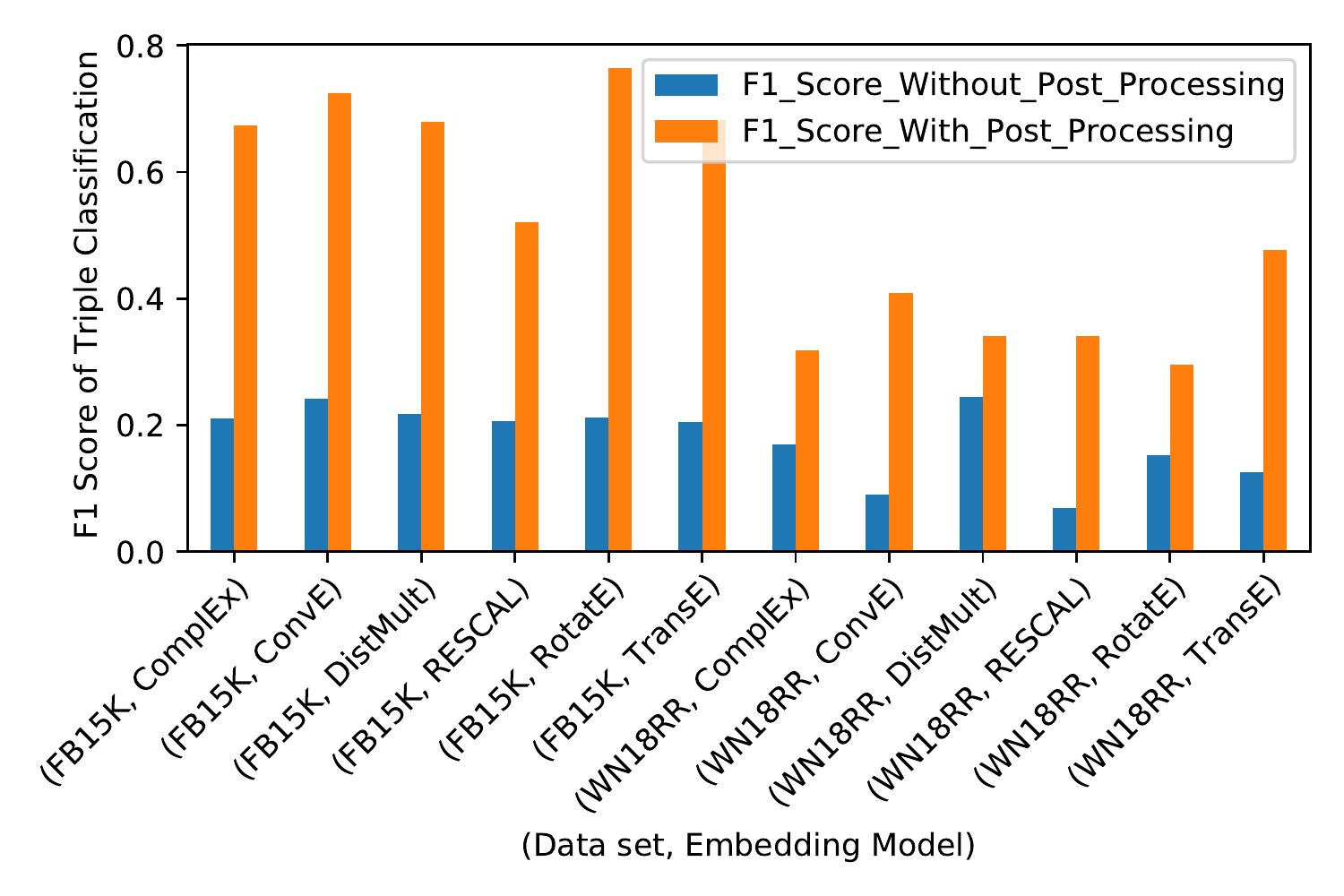}
	\caption{Comparison of Micro-F1 scores for pre-trained KGE models with and without the fine-tuning of PTSS for the triple classification task, restricted to the multi-predicate triples from each graph.}
	\label{fig:ptss_f1_scores}
\end{figure}

In addition to measuring the impact of the seed embeddings, we also analyze the lift in Micro-F1 our method provides over simply using the initialized vectors as features in the triple classification task.
We restrict the triple classification task to include only triples with multiple predicates between the head and the tail.
This allows us to asses the impact our architecture has on the problem area identified in Section~\ref{sec:introduction}, namely, the missing predicate information when aggregating the head and tail representations.
These metrics are shown in Figure~\ref{fig:ptss_f1_scores}.

In every case, we see a lift in triple classification Micro-F1 on triples with multiple predicates when using our approach.
As suspected, this increase is greater on FB15K-237, where 20\% of the graph features such triples.
Even on WN18RR, where the multi-predicate issue is less prevalent, we still see higher-quality triple embedding vectors generated from the PTSS approach.
For FB15K-237, the two best performing models are generated from using ConvE and RotatE as seed vectors, indicating that these seed approaches may have increased capacity for encoding multi-predicate information into the head and tail embeddings due to their more complex architectures and scoring functions.
This concurs with our findings in the ablation of PTSS, where ConvE and RotatE as seed embeddings to our approach outperformed their respective pre-trained models.

\subsection{Analysis of Clusterability}

We report the Calinski-Harabasz index for all triple embedding spaces in Table~\ref{tab:ch-scores}.
For both datasets, we find higher CH scores for the PTSS approach over Triple2vec.
Additionally, scores are across the board higher on FB15K-237 than WN18RR, indicating we are able to learn higher-quality vectors from this graph.
To test our hypothesis that triple vectors demonstrating greater clusterability will perform well on the triple classification task, we measure the correlations between these two scores.
There is a strong, positive correlation ($\rho = 0.7820$) between the CH metric and Micro-F1 scores on the triple classification task for Freebase.
The correlation is weaker for WN18RR, where the embeddings generated from PTSS alone have correlation of $\rho = 0.5421$ with the CH index.
When analyzing the CH scores on FB15K-237, we find that the triple embedding spaces generated using ConvE and RotatE lead to the highest scores and thus are most clusterable.
This agrees with our findings in overall triple classification performance and our ablation of the PTSS mechanism.
These two approaches lead to the highest-quality triple embedding spaces; these spaces cluster well with respect to the predicate labels, and this clusterability is a factor in performance on the triple classification task.

\begin{table*}[]
    \caption{Calinski-Harabasz scores for evaluating the quality of k-means clustering on each respective triple embedding space.}
    \label{tab:ch-scores}
    \centering
    \begin{tabular}{lcccccc}
\hline 
\multicolumn{7}{c}{WN18RR} \\
\hline 
 & d=32 & d=64 & d=128 & d=256 & d=512 & d=1024 \\
Triple2vec & 1.0809 & 1.1909 & 1.1211 & 1.0642 & 1.0904 & 1.1031 \\
\hline
Agg.	&	ComplEx	&	ConvE	&	DistMult	&	RESCAL	&	RotatE	&	TransE \\
\hline
HT	&	10.3760	&	13.1284	&	1.4678	&	11.1171	&	4.3188	&	19.5217 \\
Had.	&	\textbf{19.1498}	&	\textbf{23.7984}	&	1.9967	&	\textbf{19.2833}	&	5.0234	&	93.6061 \\
Avg.	&	9.8583	&	16.0167	&	2.4876	&	12.4854	&	\textbf{12.5828}	&	28.7761 \\
L1	&	12.3999	&	14.1780	&	\textbf{4.2113}	&	10.6682	&	9.0190	&	74.1507 \\
L2	&	15.8628	&	17.9349	&	1.7894	&	11.9538	&	2.3756	&	\textbf{88.6926} \\
\hline
\multicolumn{7}{c}{FB15K-237} \\
\hline
 & d=32 & d=64 & d=128 & d=256 & d=512 & d=1024 \\
Triple2vec & 0.9838 & 1.000 & 0.9976 & 0.9875 & 0.9924 & 0.9989 \\
\hline
Agg.	&	ComplEx	&	ConvE	&	DistMult	&	RESCAL	&	RotatE	&	TransE \\
\hline
HT	&	260.9607	&	\textbf{472.0966}	&	254.6109	&	3.0954	&	\textbf{489.9187}	&	282.7902 \\
Had.	&	\textbf{266.2164}	&	454.5830	&	\textbf{258.2619}	&	\textbf{157.2214}	&	447.6694	&	185.8585 \\
Avg.	&	255.6484	&	454.1992	&	247.7288	&	4.2279	&	454.4069	&	295.6048 \\
L1	&	261.3924	&	450.9421	&	253.9956	&	2.9730	&	454.5488	&	\textbf{298.8541} \\
L2	&	253.7461	&	459.0331	&	252.6248	&	4.6564	&	450.9278	&	181.9362 \\
\hline
    \end{tabular}
\end{table*}

%
%
\section{Related Work}\label{related}
\label{sec:related-work}

We cover the requisite background on the knowledge graph embedding models tested in our experiments.
For a full coverage of these techniques, please see~\cite{ke-survey}.
These methods fall into three main groups: translation-based models, semantic-matching models, and graph-based models.

\subsection{Translational Models}

The intuition behind translation-based models is to build vector representations of $\langle h, p, t\rangle$ such that $\emb{h} + \emb{p} \approx \emb{t}$.
Model choices then depend on which space or spaces the entities and relations are embedded in as well as the scoring function used to help the model learn to differentiate between true triples from the graph and noise triples that do not reflect real-world facts.
\textbf{TransE}~\cite{transe} is the simplest of these models.
It embeds entities and predicates by using a distance function defined by
\begin{equation}
f_p(\emb{h},\emb{t}) = -\norm{\emb{h} + \emb{p} - \emb{t}}_{1/2}
\end{equation}

While this model is simple, it struggles to properly encode one to many triples, where a single relation may hold between a head entity and several tail entities.
An entire family of translation-based models exists, adding various complexities and constraints to the embedding spaces and operations used to best recover the relations described in the original graph.

\subsection{Semantic-matching Models}

While the translational assumption $\emb{h} + \emb{p} \approx \emb{t}$ gives good geometric intuition as to the types of relations learned during model training, it is prohibitive for a wide class of relations, including those with anti-symmetric or complex properties.
Semantic-matching models deviate away from the distance-based assumption and focus on using similarity-based scoring functions to recover facts from the low-dimensional representations of entities and relations.
Rather than relying on norms and translations, these methods leverage dot-product-like scoring functions to measure angles between low-dimensional representations, sometimes referred to as `semantic energy' functions.
The simplest of such models is \textbf{RESCAL}~\cite{rescal}, which relies on a tensor representation of the underlying knowledge graph $X$, where each entry of the tensor $\emb{X}_{ijk} = 1$ if the fact is represented in the knowledge graph, otherwise zero.
This tensor can then be factorized into latent components,
\begin{center}
$\emb{X}_{k} \approx \emb{A} \emb{P}_k \emb{A}^\top$ for $k = 1, \ldots , r$
\end{center} 
where $P_k$ is a matrix of dimension $r \times r$ representing interactions between each corresponding component and $A$ contains the $r$ dimensional representations of the entities. 
Thus, for each $(h,t)$ pair, we can compute the likelihood they participate in the $k$-th relation as
\begin{center}
$f_k(\emb{h}, \emb{t}) = \emb{h}^\top \emb{P}_k \emb{t}$.
\end{center}
Contrasted with translation-based models, RESCAL takes advantage of vector products while capturing interactions between elements of each entity and all relations. 
In a simplification, \textbf{DistMult}~\cite{distmult} requires each $P_k$ to be diagonal, reducing the parameters of the model while sacrificing some of its representational capacity.
This reduction in capacity is especially felt when modeling anti-symmetric relations as interactions in these diagonal matrices have no notion of directionality.
To circumvent this issue, the \textbf{ComplEx}~\cite{complex} model allows for the low-dimensional representations to live in the complex space $\mathbb{C}$.
The scoring function used by the ComplEx model is defined as
\begin{center}
$f_k(\emb{h}, \emb{t}) = \Re(\langle \emb{h}, \emb{p}_k, \emb{t} \rangle)$ where $\emb{p}_k \in \mathbb{C}^r$.
\end{center}
By allowing the representations to be complex-valued, the model can handle the asymmetries of many relations present in knowledge graphs, yet score the likelihoods of facts existing using only the real-valued vectors.
The RotatE approach also utilizes complex-valued representations, but it also models each individual relation as a rotation between the head and tail entities.
This additional parameterization allows for increased learning capacity with the end goal of capturing anti-symmetry, inversion, and composition.
The work of~\cite{conve} takes this one step further, defining the \textbf{ConvE} model where entities interact through the convolution operator. 
This introduces additional non-linearities through which the model can increase the capacity for learning complicated relational structures.

\subsection{Graph Models}\label{sec:graph-models}

Besides exploiting the information in knowledge graph triples through translational or factorization models, graphical approaches seek to exploit information contained in the graph's structure itself.
This represents a shift from thinking of knowledge graphs embeddings of having separate embedding spaces for entities and relations.
In the work of ~\cite{node2vec}, the authors capture this shift by suggesting aggregation methods for combining nodes (entities in the KG) to represent edges (triples in the KG). 
We reproduce the summary of these operators along with new aggregation methods we use in our study in Table~\ref{tab:agg-methods}.

In order to generate node embeddings to use with these aggregation schemes, the \textbf{node2vec} approach utilizes random walks over the graph to generate sequences of nodes, where each sequence is treated as a sentence and fed into a word2vec-like model~\cite{node2vec}.
In our experiments, we consider the \textbf{Triple2vec}~\cite{triple2vec} approach, which directly learns triple representations without using aggregations of nodes by converting the original graph $G$ to a line graph $G_L$.
To reflect edge weights in this new line graph, the authors introduce the triple frequency and inverse triple frequency, defined as
$$ TF(p_i, p_j) = \log(1 + \emb{C}_{i,j}) $$
$$ ITF(p_j, E) = \log \frac{|E|}{|{p_i: \emb{C}_{i,j} > 0}|} $$
where $C_{i,j}$ counts the number of times the predicates $p_i$ and $p_j$ link the same heads and tails, and $E$
is the set of edges.
They use these metrics to build a symmetric matrix 
$$\emb{C}_M(i,j) = TF(p_i, p_j) \times ITF(p_j, E)$$
to represent a vector for each predicate in the graph; each vector in $\emb{C}_M$ can then be used to build a matrix of pairwise similarities between predicates, denoted $\emb{M}_R$.
This weighting scheme is expected to push the embeddings of triples with similar predicates to similar neighborhoods in the embedding space; we hypothesize that if this metric is effective, the resulting triple embedding vectors should serve as good features for triple classification.
Given this weighted line graph, a random walk strategy to create sequences of triples is then applied. 
The walk-based nature of this approach suits small, connected graphs and hence may not extend to larger, heterogeneous and complex-typed knowledge graphs.

\subsection{Post-Processing of Embeddings}
Aside from leveraging previous knowledge graph embedding models, our work introduces a novel way of repurposing these embeddings for alternate tasks.
This work was inspired by post-processing procedures for word and sentence embeddings, many of which are detailed in~\cite{word-pp}. 
A key idea in this area is to regularize the embedding space, leaning on the notion of isotropy of the space to better reflect pairwise similarities, which we directly estimate in an unsupervised way in our approach.
The first of these was published by~\cite{tough-to-beat}, who utilize pre-trained word embeddings, average the corresponding vectors for each word in a sentence for a singular representation, and then modify that representation using principal component analysis and singular value decomposition.
The authors find that many of the dimensions used in the sentence representation are unnecessary and subsequently introduce noise; the application of PCA/SVD reduces this noise and builds more compact features that retain the majority of the learned information.
The removal of these dominating components, some of which merely reflect the frequency of a word in the underlying corpus, leads to more robust representation.
In the work of~\cite{gem}, the authors leverage a pre-trained word2vec model (GloVe~\cite{glove} in their case).
By analyzing the geometry of the subspace generated by building a matrix $\emb{A} \in \R^{d \times n}$ for $d$-dimensional embeddings of the $n$ words in a given sentence, the authors build a new orthogonal basis vector that captures the general semantics of the words and their contexts.
Applying a sliding window QR factorization to the matrix $\emb{A}$ and re-weighting by three metrics: a word's novelty, significance, and uniqueness--a new semantically motivated sentence representation is generated. 
Not only is the approach computationally efficient, but it allows for the re-purposing of pre-trained word vectors, such as those from GloVe, allowing for new representations to be generated using no new training data.
The work of~\cite{kalinowski-sentence-structure} found that these sentence representations best align with low-dimensional knowledge graph representations; we hope to extend their work utilizing our novel triple representations.

%
%
\section{Conclusion and Future Work}\label{conclusion}
\label{sec:conclusion}

In this work, we introduce an approach for re-using pre-trained 
knowledge graph embeddings for the purpose of building explicit triple representations.
To this end, we introduce the pairwise triple similarity score as a means of bootstrapping similarity scores 
between triples in a weakly supervised way.
Using these scores, we develop a novel model to build semantically motivated triple embeddings.
We show that this model consistently beats the state-of-the-art method, Triple2vec, on two well-known knowledge graph benchmarks.
Future work on our methodology involves additional hyperparameter tuning, an ablation study on the impact of the number of pairwise triple similarity score samples and expansion to other knowledge graph benchmark datasets.
We are also interested in investigating alternatives to our PTSS scoring scheme that may better leverage information contained in the pre-trained seed embeddings, such as including the notion of triple frequency and inverse triple frequency used to calculate predicate weights in~\cite{triple2vec}.

\section*{Acknowledgment}
We would like to acknowledge the authors of Triple2vec~\cite{triple2vec}, who shared their implementation to serve as a baseline for comparison. 
We would also like to thank the maintainers of LibKGE~\cite{libkge} for providing the pre-trained knowledge graph embedding models.

\bibliographystyle{ieeetr}
\bibliography{semanticstructure}

\end{document}